\documentclass{ifacconf}

\usepackage{array}
\usepackage{amssymb}
\usepackage{amsmath}
\usepackage{makecell}
\usepackage{transparent}
\usepackage{color}
\usepackage{subcaption}
\usepackage{caption}
\usepackage{xcolor}
\usepackage{tabu}

\usepackage{import}

\usepackage{graphicx}      
\usepackage{natbib}        
\begin{document}
\begin{frontmatter}

\title{Robust Data-Driven Error Compensation for a Battery Model} 



\author[First]{Philipp Gesner}, 
\author[First]{Frank Kirschbaum},
\author[First]{Richard Jakobi},
\author[Second]{Ivo Horstkötter},
\author[Second]{Bernard Bäker}

\address[First]{Mercedes-Benz AG, Mercedesstr. 137, 70327 Stuttgart}
\address[Second]{Institute of Automotive Technology Dresden, George-Bähr-Str. 1, 01069 Dresden}

\begin{abstract}                
Models of traction batteries are an essential tool throughout the development of automotive drivetrains. Surprisingly, today's massively collected battery data is not yet used for  more accurate and reliable simulations.
Primarily, the non-uniform excitation during regular battery operations prevent a consequent utilization of such measurements. Hence, there is a need for methods which enable robust models based on large datasets. \\
For that reason, a data-driven error model is introduced enhancing an existing physically motivated model. A neural network compensates the existing dynamic error and is further limited based on a description of the underlying data. This paper tries to verify the effectiveness and robustness of the general setup and additionally evaluates a one-class support vector machine as the proposed model for the training data distribution.
Based on a five datasets it is shown, that gradually limiting the data-driven error compensation outside the boundary leads to a similar improvement and an increased overall robustness.
\end{abstract}


\end{frontmatter}

\section{Introduction}
\subsection{Motivation}
Batteries play a dominant role in the electrification of propulsion systems. Consequently, characterizing and modeling the performance of lithium-ion batteries is among the higher priorities of automotive manufacturers. The typical engineering models used in this context are based on analytical findings of the battery cells. There is also emerging research which extends existing physical models with purely data-driven models. Such approaches are especially useful in later development phases, like the targeted application of a model-based battery simulator powering electrical axles for testing purposes.\\
\cite{Ljung.2008} already emphasized the significance of utilizing today's extensively collected data for system modeling. Among other reasons, the inhomogeneous and partial excitation of the system is still stopping a widespread use of large datasets. \\
A very promising concept to tackle the subsequent challenges is to combine analytical models with more data-driven approaches. This idea potentially connects excellent extrapolation capabilities of physically motivated models with high accuracy of black-box models. \\
Thus, this paper aims at an effective methodology, which enhances physical models with data-driven approaches while still ensuring a robust output. This is accomplished by limiting the influence of the data-driven error compensation based on a boundary description of the training data. The fundamental assumption behind this is that a black-box model can only generalize to a certain degree. Accordingly, a \textit{one-class support vector machine}, as the boundary model is evaluated regarding an increase in robustness of the overall model structure. Robustness in this context is understood as a reliable and stable model performance independent of the present inputs. Ultimately, this work tries to build more trust in regard to data-driven approaches based on a previous study (\cite{Gesner.2020}).
Firstly, this paper presents related work with a focus on modeling data distributions and then introduces an example supporting this fundamental idea. Afterwards the methods regarding the error compensation structure and the applied boundary model are laid out. Lastly, the paper quantifies the results on particular validation datasets and concludes the findings.
\subsection{Related work}
Modeling the behavior of automotive batteries usually relies on understanding the electrochemical phenomena. Most model structures found in the literature capture basic effects like diffusion and double-layer capacitance by using \textit{equivalent circuit models} (e.g. \cite{Birkl.2013b}). Some researchers, for example \cite{Buller.2003}, built even more detailed models integrating additional aspects of the voltage response like the Butler-Volmer relation. Data-driven models have been applied to capture the nonlinear battery dynamics as well, e.g.  by \cite{Capizzi.2011} using \textit{recurrent neural nets}. \\
Surprisingly, there are no recognizable efforts to obtain accurate and robust models of the battery dynamics based on extensive measurements, as they are produced in vehicle development process. In general, studies related to nonlinear dynamic modeling still lack efficient algorithms to handle large amounts of data.\\
Mining and understanding important information within large datasets plays a more significant role in the \textit{machine learning} community. In this sense, \cite{Nelles.2001} suggests hybrid model structures, that are composed of submodels consisting of a first principle model and a submodel capturing missing phenomena by using data-driven approaches. Three basic hybrid frameworks can be distinguished. For instance, \cite{Bikmukhametov.2020} present a \textit{parallel hybrid model}, which is used to compensate the prior output error of a physical model with a multilayer neural network. A \textit{series hybrid model} manipulates the in- and outputs of the physical model with data-driven methods. The \textit{parameter scheduling hybrid model} is a more common method to enhance first principle models with measurement data. For example, \cite{Psichogios.1992} use a feedforward neural net to estimate the process parameters of a fedbatch bioreactor model.\\
There are also models that combine the two modeling approaches in regard to the electric battery behavior. \cite{Park.24.05.201726.05.2017} introduce an \textit{Elman network}, which not only compensates the error of an electrochemical model, but also captures missing dynamics.\\
One challenge, that is unavoidable when modeling purely relying on data, is the weakness in regard to extrapolation. However, this can be tackled with a hybrid model. The hybrid structure allows the reduction of the influence of the data-driven model whenever it is not interpolating. As a result, describing the boundaries of the underlying data distribution is fundamental. \\
This task is well known in the field of model-based optimization due to the importance of a reliable model output. There are variety of  boundary models especially in the context of engine calibration. \cite{Renninger.2005} for instance use a convex hull to describe the boundaries of their design space. Similar approaches are applied for detecting anomalies, for example credit card fraud. \cite{Hodge.2004} describe how the boundaries of a dataset can be described by stochastic or density-based methods like, e.g. \textit{k-nearest neighbor}. Regularly, the \textit{support vector machine}, a popular classification tool, is used to support engineering models (\cite{Kampmann.2012}).\\
\textit{Generative models } are a trending method for obtaining new samples of a given dataset.  Therefore, they need to model the underlying probability distribution of data as well. \cite{Sensoy.2020} shows how \textit{variational autoencoder} can detect and generate out-of-distribution samples targeting robustness in terms of adversarial examples. \\ 
None of the identified publications use boundary description of the training data to actively combine data-driven models with analytical models. Hence, this work targets a robust and safe enhancement of a physically motivated battery model based on randomly collected time series data. 
\section{Basic Idea}\label{ch:example}
\begin{figure}[h!]
    \centering
    \begin{center}
        \includegraphics[width=0.5\textwidth]{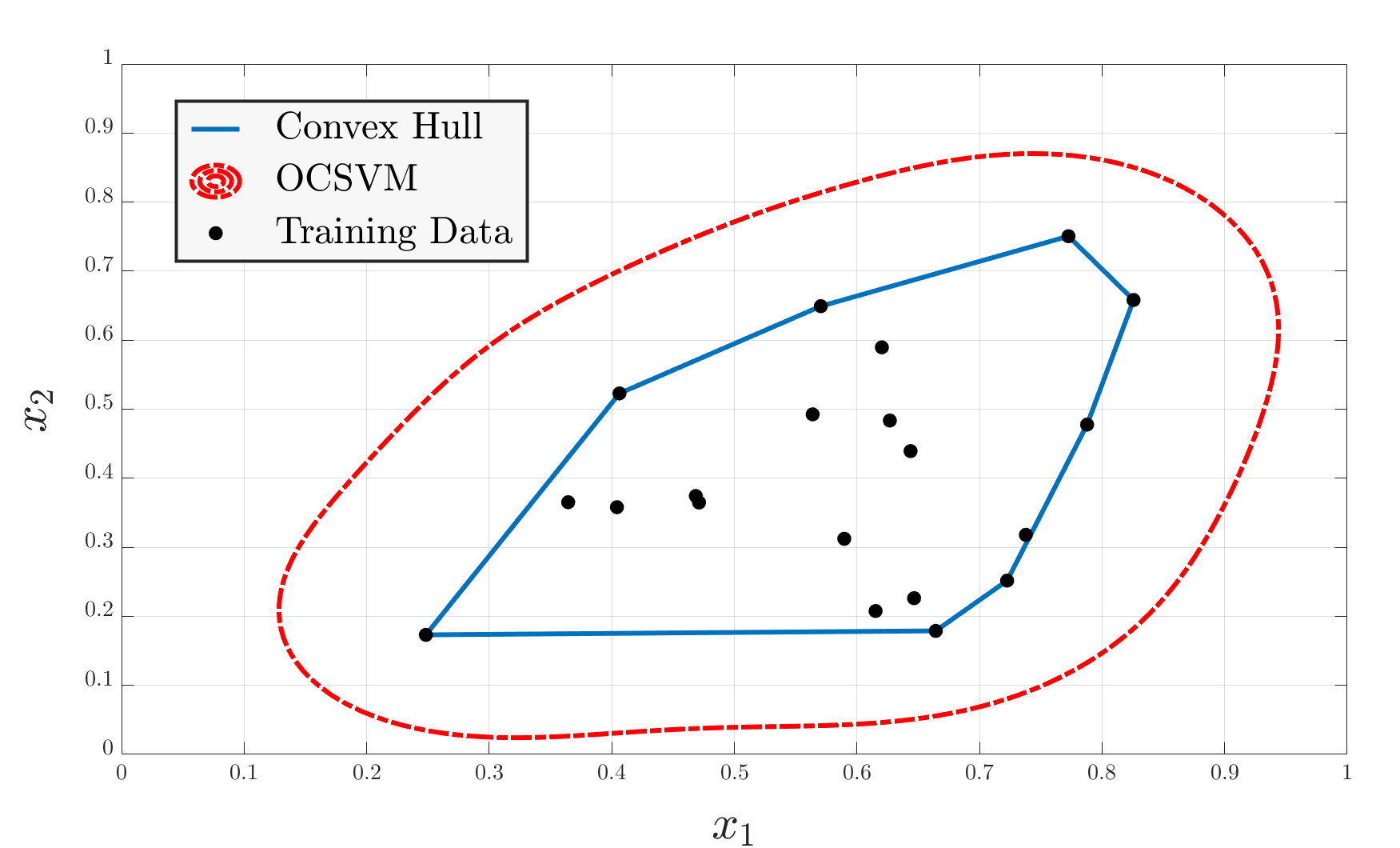}    
        \caption{Training data with boundary functions}
        \label{fig:Example}
    \end{center}
\end{figure}

This introductory example reinforces the idea that knowledge about the training data distribution can be used to prevent a purely data-driven model from false generalizations.\\
Ten random polynomial functions, based on a generator proposed by \cite{Belz.02.09.201504.09.2015}, serve as typical regression problems. The mappings are targeted to imitate engineering applications. All functions used in this example have 30 polynomial terms and an average exponent magnitude of 5. \\
Additionally, the example tries to mimic the stated problem of error compensation by incorporating sine functions into every term of the polynomials. This results in a more oscillating output $y$ with a center close to zero,  which is similar to the error of the analytical model used in later chapters. In accordance to the stated goals, the data-driven error compensation is assumed to perform better when the error model is limited to the boundaries of the training data. In other words, the method exclusively trusts the analytical model whenever the inputs strongly differ from the previously known data distribution.\\
Figure \ref{fig:Example} shows such a two-dimensional input space represented by the variables $x_1$ and $x_2$. The 20 data points were generated randomly.\\
They serve to train a \textit{feedforwad neural net} (FNN) with three layers and ten neurons. For each of the polynomial functions a different model is obtained. Weights and biases are determined based on the Levenberg-Marquardt algorithm and a random initialization.\\
The figure further illustrates two fundamental methods on how to describe the boundaries of the available data. The convex hull based on the \textit{quick hull} algorithm (\cite{Barber.1996}) simply finds a minimum number of convex lines surrounding the two-dimensional data points. Another boundary model is the \textit{one-class support vector machine} (OCSVM). \cite{Scholkopf.1999} introduced this semi-supervised classifier as a tool for detecting novelty within data. It is highly parameterizable (cf. chapter \ref{ch:boundary}) and thereby finds a more intuitive description. Among other advantages over the convex hull, it allows to model concave bounds and even holes in case there is a significant gap between different parts of the data.\\
Both models can determine whether a new data point $x$ is inside or outside the boundary. This information can then be used to limit the output $\hat{y}$ of the neural net as follows:\\
\begin{equation}
    \hat{y}(x)= 
\begin{cases}
    f_{\mathrm{FNN}}(x),& \text{if } x  \text{ is inside OCSVM}\\
    0,               & \text{if } x  \text{ is outside OCSVM}
\end{cases}
\label{bound_0}
\end{equation}
To make this dataset more realistic \textit{white Gaussian noise} is added to the in- and outputs of the training data. A signal-to-noise ratio of $40\, \mathrm{dB}$ is applied. This seems to be suitable in the context of the vehicle signals used in the following chapters. For instance the quantization and the nondeterministic bus system lead to quite noisy data, which is nonnegligible when testing new modeling methods.\\
To evaluate the approach stated in (\ref{bound_0}), a test data set containing 20,000 points is created for every polynomial function. They cover the entire hypercube $[0,\: 1] \in \mathbb{R}^2$, which is  around three times the area of the convex hull in all ten cases. The performance of the ten FNNs in regard to their validation data can be quantified with a \textit{root mean squared error} RMSE of 2.7690. When the boundary condition based on the OCSVM is applied, an average RMSE of 1.495 is determined. Hence, the condition in (\ref{bound_0}) improves the model by $78.7\,\%$, clearly supporting the stated hypothesis about the benefits of limiting a data-driven model to previously known regions of its input space.\\
This introductory example emphasizes the need of research, when it comes to modeling with heterogeneous, large datasets. \\
Usually it is assumed, that models can generalize unlimitedly. However, it can be shown that the FNNs extrapolate accurately only to a certain degree. If the narrower convex hull is used, the RMSE rises to 1.7361, because it limits output too early.\\
Undoubtedly, factors like amount of noise, number of input data points and system order have a huge influence on the generalization capabilities of the black-box model. This is due to the training of the FNN. Its weights $\omega$ and biases $b$ are determined on a comparison of the measured outputs $y(k)$ with the modeled outputs $\hat{y}(k)$:
\begin{equation}
\min_{\omega,b}\sum_{k=1}^{n}(y(k)-\hat{y}(k,\omega,b))^2
\label{eq:guetefunktional-ls-summe}
\end{equation}The more ambiguous information is used during the training, the more the model tends to overfit the data. As a result, the performance decreases in regard to underrepresented areas of the input space. This phenomenon of overfitting is well known and can be dealt with by using validation data. Accordingly, the generalization error, which is highly related to the concept of overfitting, can be reduced as well. Nevertheless, this is true only to a certain level. If there is data that is just unlike the information used to train the neural net (cf. eq. (\ref{eq:guetefunktional-ls-summe})), it can not predict accurately. \\
This basic idea is adopted to a model of a traction battery, which requires incorporating dynamic behavior, a higher dimensional input space and more severe heterogeneity of the data. In a more general understanding, this paper shows a method on how to utilize massively collected data sets on engineering systems, an aspect that needs to be addressed more in the field of system identification. 
\section{Error Compensation}
\label{ch:EC}
\begin{figure}[h!]
    \centering
    \begin{center}
        \includegraphics[width=0.46\textwidth]{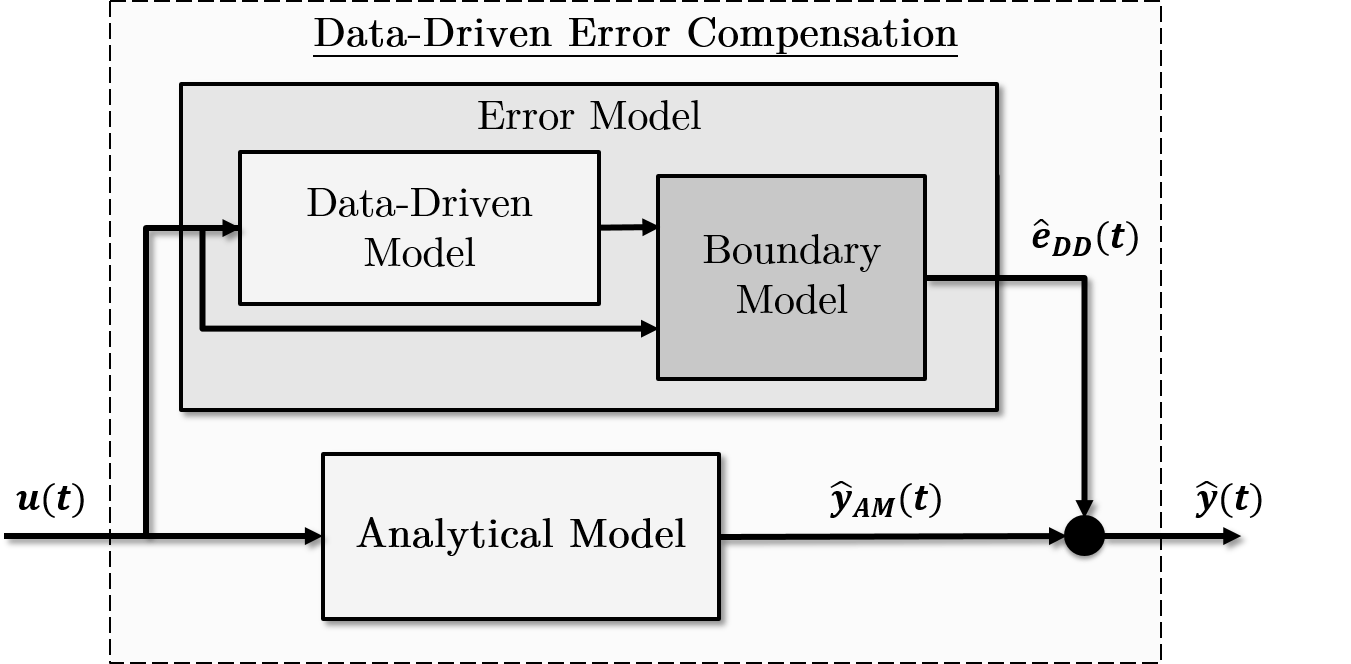}    
        \caption{Structure of the data-driven error compensation}
        \label{fig:ECM}
    \end{center}
\end{figure}

\cite{Nelles.2001} states that state-of-the-art models normally do not accurately reflect the characteristics of the underlying processes. In many cases such existing models are linear or the modeled nonlinearities are just erroneous. This motivates an enhancement of typically used analytical models by combining them with submodels, which are obtained purely from measurements. This potentially increases extrapolation capabilities, robustness and industrial acceptance in comparison to a data-driven model alone.\\
The idea of combining a prior model with a data-driven approach is the fundamental concept behind the error compensation introduced in this paper. Its structure, visualized in figure \ref{fig:ECM}, might also be understood as a hybrid parallel model. 
\\The data-driven model is integrated in such a generic way, that any existing model can be used. Typically, the measurable error $e_{\mathrm{DD}}$ of an analytical model is compensated. Both outputs are then added together leading to an improvement of the overall output $\hat{y}$.
Many of the stated advantages are applicable to this particular type of hybrid model. Most importantly and at the core of this study, the output value of the error model $\hat{e}_{\mathrm{DD}}$ can be bound easily to its interpolation capabilities. Therefore, the boundary model decides based on the input data $u(t)$, whether the compensation can be trusted or the model should only rely on the output of the analytical submodel $\hat{y}_{\mathrm{AM}}$. As a result, it can be ensured that the error compensation works in a robust manner. This translates into automatically switching between an increased data-driven accuracy and a guaranteed extrapolation stability based on the prior data distribution. \\
A more formal representation of the error compensation structure is given here:
\begin{equation}\label{ECM}
\hat{y}(t)= \hat{y}_{\mathrm{AM}}(u(t)) + g_{\mathrm{BM}}(f_{\mathrm{DR}}(u(t)),u(t))
\end{equation}
The boundary model $g_{\mathrm{BM}}$ (cf. chapter \ref{ch:boundary}) is a function that limits the output of the data-driven regression $f_{\mathrm{DR}}$. Their joined output $\hat{e}_{\mathrm{DD}}$ is then added to the prediction of the analytical model. \\
In contrast to most other concepts of combining different modeling approaches, this structure further allows to capture dynamic processes that are missing in the analytical model. \\
Thus, a discussion on the dynamics of the electric battery behavior is required. The nonlinear dynamics of the battery's voltage, which are heavily defined by the single electrochemical cells, are usually modeled based on differential equations of high orders. However, the approximation of the voltage response $y$ with a 4$^{\mathrm{th}}$ order linear system in one operational point is known to be adequate for many applications. Higher order differential equations usually indicate a pole-zero cancellation. For building battery models with a high quality, it is not enough to consider the current $u_1(t)$ as input, but also the operational point regarding the temperature $u_2(t)$ and state of charge $u_3(t)$. When it comes to the voltage output, \cite{Gesner.2019} further investigated that the battery dynamics of purely electric vehicles should be modeled up to at least $10\,\mathrm{Hz}$.\\
The prior model used for this work is a commonly used equivalent circuit with an internal resistance and two RC-elements. With the 2$^{\mathrm{nd}}$ order dynamic model the dominant processes within the cell are captured. One of the RC-elemtents normally covers the double-layer capacitance with a relatively low time constant and the other one represents the diffusion of lithium-ions with a high time constant. All parameters of the analytical model are a function of temperature, state of charge and current. Those nonlinear functions are either realized as look-up tables or physical relations like the Arrhenius equation regarding the temperature.  Furthermore, the open circuit voltage of the battery is modeled based on the measured relation between the state of charge and voltage.\\
This neglects the temperature and the time-variant hysteresis effect of the open circuit voltage. Additionally, the model  lacks an accurate description of the low frequency processes (\cite{Oldenburger.2019}).  Another downside of the given analytical model is that it does not take any inhomogeneities of the different cells into account nor does it consider any electric effects occuring on a battery system level. \\
Since the data-driven model is set out to include dynamic behavior as well a recurrent model structure, using past values of the in- and outputs is required. Besides compensating the stationary behavior, the focus of the error model is correcting the battery's diffusion processes with large time constants.  In combination with the analytical model a 1$^{\mathrm{st}}$ order nonlinear dynamic model seems sufficient, further allowing a reduction input space's dimensions.\\
For the data-driven error model a \textit{recurrent neural network} is used. The recurrence of the dynamic regression takes place outside the neural network. Such a model with no inner loops is also known as a \textit{nonlinear autoregressive exogenous model} (NARX). In terms of a dynamic system, this is an intuitive structure due to its similarities with a difference equation. Ultimately, this NARX structure increases the interpretability of the model order, enables the use of different nonlinear functions, and requires a lower training effort than other recurrent models. \\
A simple \textit{feedforward neural net} (FNN) was chosen for the regression, mainly because of its well established algorithms when it comes to large amounts of observations. It consists of three layers and the hyperbolic tangent as activation function. According to the universal approximation theorem (\cite{Hornik.1989}) one hidden layer already enables to model any arbitrary function. To determine an optimum of the cost function (cf. eq. (\ref{eq:guetefunktional-ls-summe})) the Levenberg-Marquardt algorithm is applied. Further, a stopping criterion was implemented, that depends on a change of the remaining error over epochs. If its value stays within a band of $10^{-8}$ over ten iterations, the training will terminate.\\
Before the training, the available data needs to run through the analytical model generating the error $\hat{e}_{\mathrm{DD}}(t) = \hat{y}(t) - \hat{y}_{\mathrm{AM}}(u(t))$  in accordance to equation (\ref{ECM}). Afterwards an anti-aliasing filter excluding all frequencies above $10\, \mathrm{Hz}$ is applied. The battery signals are then downsampled to $20\, \mathrm{Hz}$. For every discrete time step $k$, there are the inputs measured current $u_1(k)$, the estimated temperature $\hat{u}_2(k)$ from a thermal model, the state of charge $u_3(k)$ from a current integration and the previous values $u_1(k-1)$ and $\hat{e}_{\mathrm{DD}}(k-1)$. The neural network then approximates the present error $\hat{e}_{\mathrm{DD}}(k)$ based on the nonlinear function:
\begin{align}\label{eq:FNN}
 \,   f_{\mathrm{DR}}(u_1(k), u_1(k-1), \hat{u}_2(k), u_3(k), e_{\mathrm{DD}}(k-1))&
\end{align}
This formulation of the model as a one-step predictor is also known as \textit{series-parallel setup}. It can be used to train the FNN in the same way as the model in chapter \ref{ch:example}. However, with a simulation focus in mind, the model relies on the feedback of previously predicted outputs $\hat{e}_{\mathrm{DD}}(k-1)$. As a result, it is operated in a \textit{parallel setup} and needs to be trained specifically for a robust simulation of the dynamics. For this purpose, \cite{Williams.1995} introduced the \textit{real-time recurrent learning algorithm} (RTRL) which unrolls the FNN over the given sequential data. The RTRL is used to obtain the neural net in this paper ensuring a focus on its simulation capabilities.\\
Additionally, the number of neurons is determined based on a grid search. It is assumed that the hyperparameter tuning can be conducted on a small space-filling subset of the data (\cite{Gesner.2020b}). The FNN is trained multiple times determining the ideal number of neurons between 11 and 59. Afterwards, a final error model is trained on the entire dataset. This method consistently leads to accurate error models.\\
After the training, the error model is combined with the analytical model. Their connection is controlled by a boundary model, that ensures a safe error compensation of the analytical model.
\section{Boundary Model} \label{ch:boundary}
In the introductory example (chapter \ref{ch:example}), a boundary model of the training data improves the overall performance of the regression. Fair knowledge of the data distribution used for training is in any case useful. Even if a model generalizes well,  the prediction of nonlinear processes, that differ from the prior knowledge, is only possible to a limited extent. In addition, this statement is even more important for data-driven dynamical models (cf. eq. (\ref{bound_0})) because erroneous outputs $\hat{e}_{\mathrm{DD}}(k-1)$ are included in the input space and can potentially cause unstable simulations.\\ 
For the boundary function $g_{BM}$ the \textit{one-class support vector machine} (OCSVM) is chosen. It is a promising tool in the field of anomaly and novelty detection.
Moreover it is an adaption of the \textit{support vector machine}, which is a classification method that can work only with few data points. \cite{Scholkopf.1999} introduced OCSVM by removing labels and establishing a new hyperparameter $ \nu \in (0,1) $, which limits the Lagrange multipliers $ \alpha $ in combination with the number of training data points $l$.  This results in a quadratic optimization problem for the OCSVM: \\
\begin{align}\label{DualQP-OCSVM}
\min_{\alpha} \quad \: \: \: \: & \dfrac{1}{2} \: \sum_{i=1}^l \sum_{j=1}^l \alpha_{\mathrm{i}} \alpha_{\mathrm{j}} \: K(u_{\mathrm{i}}, u_{\mathrm{j}}) 
\\
s.t.\qquad & \sum_{i=1}^l \alpha_{\mathrm{i}} = 0 
\\
& 0 \leq \alpha_{\mathrm{i}} \leq \frac{1}{\nu l} 
\end{align}
For the kernel function $K(u_{\mathrm{i}}, u_{\mathrm{j}}) $ of the data points $u(k)$ the Gaussian kernel is selected. It is related to the Gaussian normal distribution and also includes an adjustable parameter $ \sigma $. Together with $ \nu $ two hyperparameters are identified, that have a significant influence on the model. \\
Finding the optimal Lagrange multipliers $ \alpha $ results in classifier according to the following equation:
\begin{equation} \label{SVM_Klassifikator_OCSVM}
f_{\mathrm{OC}}(u(k)) = (\sum_{i=1}^l \alpha_{\mathrm{i}}  \: K(u(k), u_{\mathrm{SV}}) - b)
\end{equation}
The function $f_{\mathrm{OC}}(u(k))$ outputs the distance of any data point $u(k)$ to the hyperplane defined by the support vectors $u_{\mathrm{SV}}$. Its sign defines whether the input vector belongs to the learned class (positive values) or lies outside the hyperplane (negative values). \\
The kernel parameter $\sigma$ and the regulation parameter $\nu$ are tuned based on the confusion matrix for classification. Therefore the OCSVM is compared with the convex as a baseline for the classifier. In this sense, the False Positive Rate (FPR) assesses the degree to which the hyperplane of the OCSVM intrudes into the areas outside the convex hull. Additionally, the False Negative Rate (FNR) calculates the proportion of incorrectly classified regions within the convex hull. Both criteria are used to conduct a grid search on the hyperparemeters.\\
With the bias $b$ in equation (\ref{SVM_Klassifikator_OCSVM}) it is further possible to make the classifier's boundary narrower or wider. In figure \ref{fig:Example} the bias is increased compared to the value the optimization algorithm determines. Thereby it allows a certain degree of extrapolating the training data, which results in a higher performance than limiting the regression model based on the narrower convex hull.  Thus, the data-driven model is usually capable of generalizing to some degree.\\
For the interplay between the boundary model $f_{\mathrm{OC}}(u(k))$ and the data-driven regression $f_{\mathrm{DR}}(u(k))$ it is proposed to use a function that gradually decreases the influence of the error model.  Similar to the condition in (\ref{bound_0}) the introduced boundary model outputs the following error compensation $\hat{e}_{\mathrm{DD}}(u(k))$:

\begin{equation}
=
	    \begin{cases}
   f_{\mathrm{DR}}, & \text{if  }  f_{\mathrm{OC}}   > 0\\
  	f_{\mathrm{DR}} \, \mathrm{sig}(-\gamma \, f_{\mathrm{OC}}) ,            & \text{if  } f_{\mathrm{OC}}  \leq 0
\end{cases}
\label{bound_DR}
\end{equation}

The parameter $\gamma$ enables to adjust the steepness of the limitation. It is set to 2 for this work, but should be part of a broader algorithm for optimizing the interplay. \\
Other functions than the proposed $sigmoid()$ might be used in the general structure (cf. figure \ref{fig:ECM}). Instead of finding the perfect modeling method, this paper rather tries to evaluate, if a limitation of the error compensation is beneficial.
\section{Validation Approach and Results}
For this study almost 142\,h of driving data are selected. Applying the downsampling leads to 10,224,000 discrete time steps to train and validate the model. As it is natural for real driving, much of the available battery data is concentrated near the system's equilibrium. The introductory example in mind (chapter \ref{ch:example}), there are sparse areas of the training data distribution, that need to be known and consequently be used to limit the regression of the NARX model.
\begin{figure}[h!]
    \centering
    \begin{center}
      \includegraphics[width=0.48\textwidth]{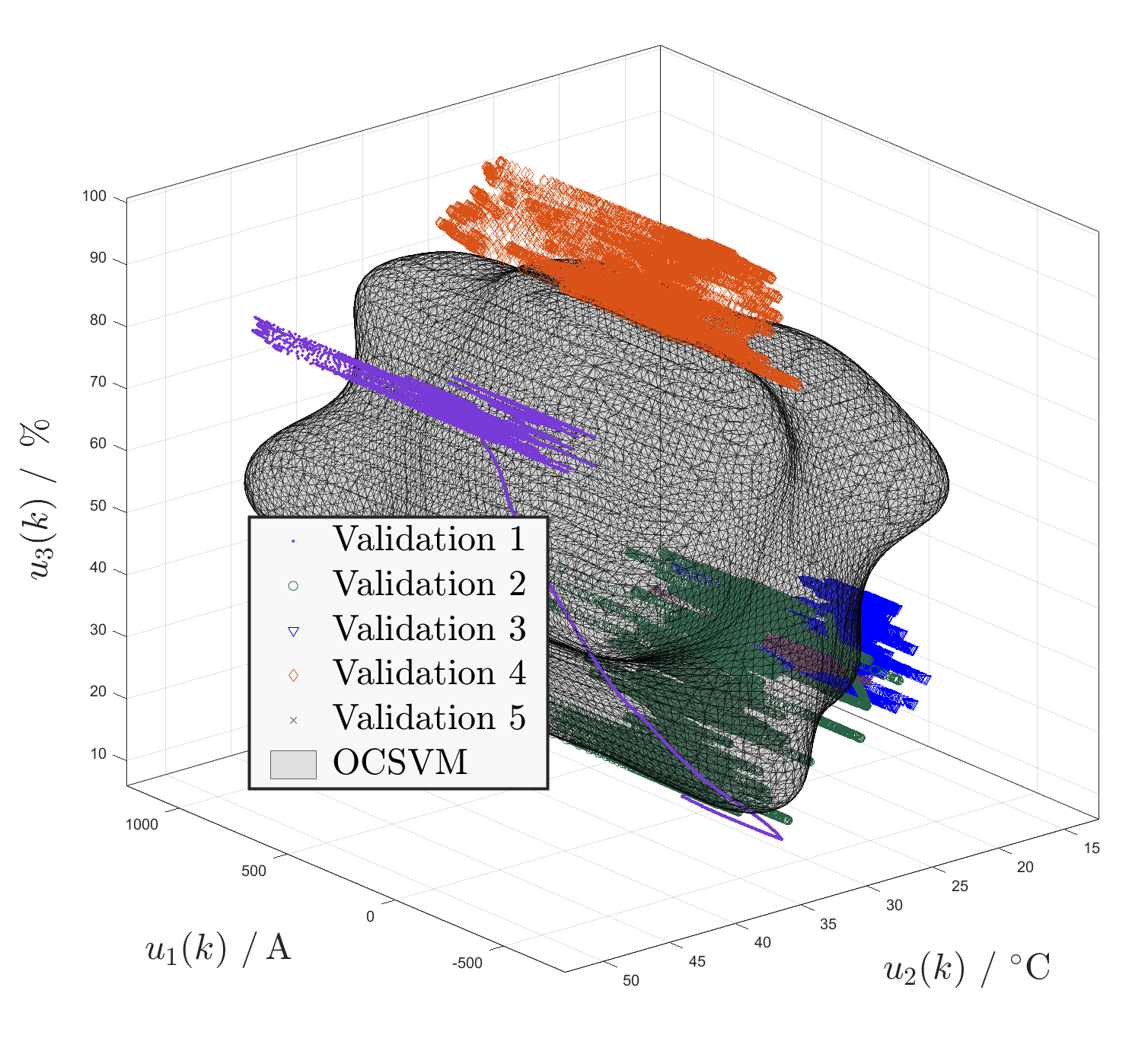}    
        \caption{Validation datasets and OCSVM}
        \label{fig:Validation_data}
    \end{center}
\end{figure}
\\For a better validation of the proposed robust error compensation model ($\mathrm{ECM}$), five driving cycles were selected that are near the edge of the training data. Figure \ref{fig:Validation_data} shows this data via a projection onto the input subspace of current $u_1(k)$, temperature $u_2(k)$, and state of charge $u_3(k)$. More precisely, the datasets were selected based on the portion of their data points that are inside the convex hull (0\,\%, 29.8\,\%, 34.1\,\%, 50.5\,\% and 67\,\%).\\
Additionally, three different models are trained on the remaining data ensuring reproducible results. On average, it takes $3.4\,\mathrm{h}$ to train one of the neural networks. The OCSVM is conducted within 13$\,\mathrm{min}$ based on a space-filling subset of the training data.
\begin{figure}[h]
    \centering
    \begin{center}
        \includegraphics[width=0.5\textwidth]{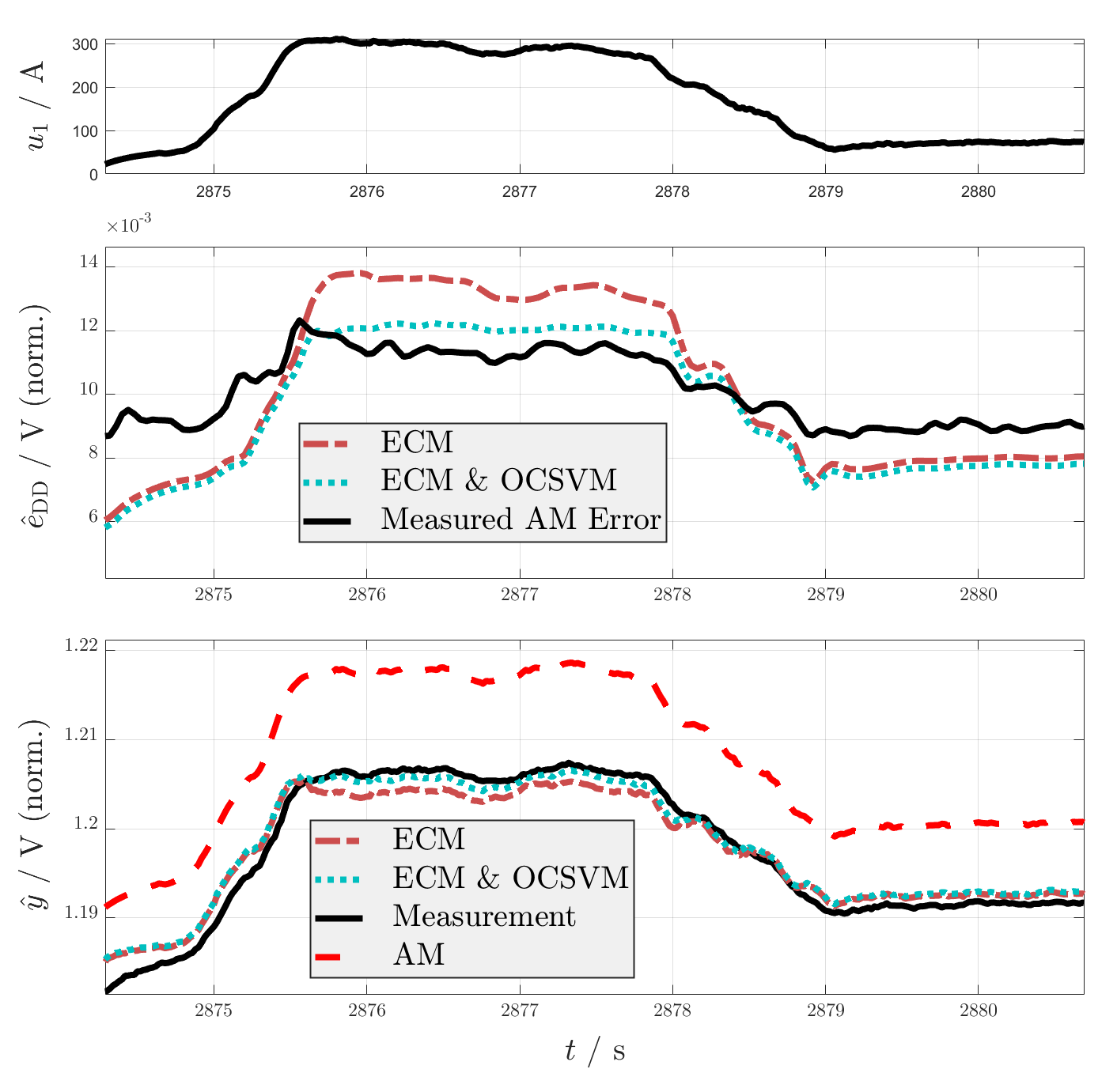}    
        \caption{ECM performance on Validation 3}
        \label{fig:Validation}
    \end{center}
\end{figure}
\\The performance of the introduced approach (cf. chapter \ref{ch:EC} and \ref{ch:boundary}) is validated on those five datasets. Figure \ref{fig:Validation} shows the input $u_1$, the normalized output of the error model $\hat{e}_{\mathrm{DD}}$ and the normalized overall output $\hat{y}$ of the hybrid model regarding the validation dataset with 34.1\,\% data points inside the convex hull. The second plot shows that the error of the analytical model (AM) is fit well by the data-driven model. When the boundary model is added to the data-driven regression (ECM $\&$ OCSVM), there are cases where the compensation is trusted less. In the first quarter of the plot this happens in an ideal manner causing a more accurate voltage output $y$.\\
Quantifying the results leads to the mean errors in table \ref{tb:margins}. The comparison now includes a limitation based on the convex hull as well (ECM $\&$ Convex Hull). In contrast to figure \ref{fig:Validation}, the OCSVM is not improving the ECM in terms of the \textit{root mean squared error} (RMSE). Nonetheless, it reduces the maximum error, which can be interpreted as an successful limitation of the data-driven submodel. On average, the introduced approach reduces large errors preventing the compensation model to generalize erroneously.  The OCSVM furthermore limits the error model in a smoother and more effective way than the convex hull.\\
\begin{table}[h!]
    \begin{center}
        \caption{Mean validation performance of ECM }\label{tb:margins}
        \taburulecolor{grey}
        \begin{tabular}{lcc}
	            & $\overline{RMSE}$  &  $\overline{Max.\:Error}$ (norm.) \\ \hline
           AM & 0.56 &   0.0159 \\ \hline
           ECM  & 0.37 &  0.0167 \\
           	ECM \& OCSVM & 0.38 & 0.0152\\ 
            ECM \&  Convex Hull  & 0.51 &  0.0158   \\ \hline
        \end{tabular}
    \end{center}
\end{table}
\section{Conclusion}
There is a clear advantage of the data-driven error compensation in regard to accuracy. Additionally, the introduced boundary model for ensuring a safe and robust operation of dynamic neural networks is clearly beneficial in terms of large errors. The \textit{one-class support vector machine} successfully reduces the influence of the data-driven submodel in accordance with the underlying data distribution.  This ultimately increases the trust on data-driven models, especially in the context of engineering applications like physical simulators. \\
Nonetheless, the proposed hybrid model still needs be tested under more conditions. Future research additionally targets modeling time-variant battery effects like ageing and hysteresis. 
\bibliography{SYSID}             

\end{document}